\documentclass[10pt,twocolumn,letterpaper]{article}
\usepackage{cvpr}
\usepackage{cvpr24}
\usepackage{tabularx}
\def\paperID{7308}
\def\confName{CVPR}
\def\confYear{2024}
\newcommand{\dataset}[0]{\acs{dataset}\xspace}
\newcommand{\datasetfull}[0]{\aclu{dataset}\xspace}

\newcommand{\mocap}[0]{MoCap\xspace}
\newcommand{\vicon}[0]{VICON\xspace}

\newcommand{\articulated}[0]{{\color{pink}[a]}\xspace}
\newcommand{\rigid}[0]{{\color{blue!40}[r]}\xspace}

\acrodef{hsi}[HSI]{Human-Scene Interaction}
\acrodef{ik}[IK]{Inverse Kinematic}
\acrodef{hoi}[HOI]{Human-Object Interaction}
\acrodef{dataset}[\texttt{TRUMANS}]{\underline{T}racking H\underline{um}an \underline{A}ctio\underline{n}s in \underline{S}cenes}
\acrodef{ahoi}[AHOI]{Articulated Human-Object Interaction}
\acrodef{fahoi}[f-AHOI]{Full-Body Articulated Human-Object Interaction}
\acrodef{cvae}[cVAE]{conditional Variational Auto-Encoder}
\acrodef{icp}[ICP]{Iterative Closest Points}
\acrodef{tlcc}[TLCC]{time-lagged cross-correlation}
\acrodef{urdf}[URDF]{Unified Robot Description Format}

\makeatletter
\renewcommand{\paragraph}{%
  \@startsection{paragraph}{4}%
  {\z@}{1ex \@plus 1ex \@minus .2ex}{-1em}%
  {\normalfont\normalsize\bfseries}%
}

\newcolumntype{C}{>{\centering\arraybackslash}X}
\makeatother

\title{Scaling Up Dynamic Human-Scene Interaction Modeling\vspace{-9pt}}

\begin{document}

\author{%
    Nan Jiang$^{1,2\,*}$, Zhiyuan Zhang$^{1,2\,*}$, Hongjie Li$^{1}$, Xiaoxuan Ma$^{3}$, Zan Wang$^{4}$, \\Yixin Chen$^{2}$, Tengyu Liu$^{2}$, Yixin Zhu$^{1\,\textrm{\Letter}}$, Siyuan Huang$^{2\,\textrm{\Letter}}$
    \vspace{3pt}\\
    \small $^1$Institute for AI, Peking University \quad
    \small $^2$National Key Lab of General AI, BIGAI \quad
    \small $^3$School of Computer Science, CFCS, Peking University\\
    \small $^4$Beijing Institute of Technology \quad $^\star{}$Equal contributors \quad $\textrm{\Letter}$\,\,\texttt{yixin.zhu@pku.edu.cn,\,syhuang@bigai.ai}
    \vspace{3pt}\\
    \href{https://jnnan.github.io/trumans/}{https://jnnan.github.io/trumans/}
    \vspace{-21pt}
}

\twocolumn[{%
    \renewcommand\twocolumn[1][]{#1}%
    \maketitle
    \begin{center}
        \centering
        \captionsetup{type=figure}
        \includegraphics[width=\linewidth]{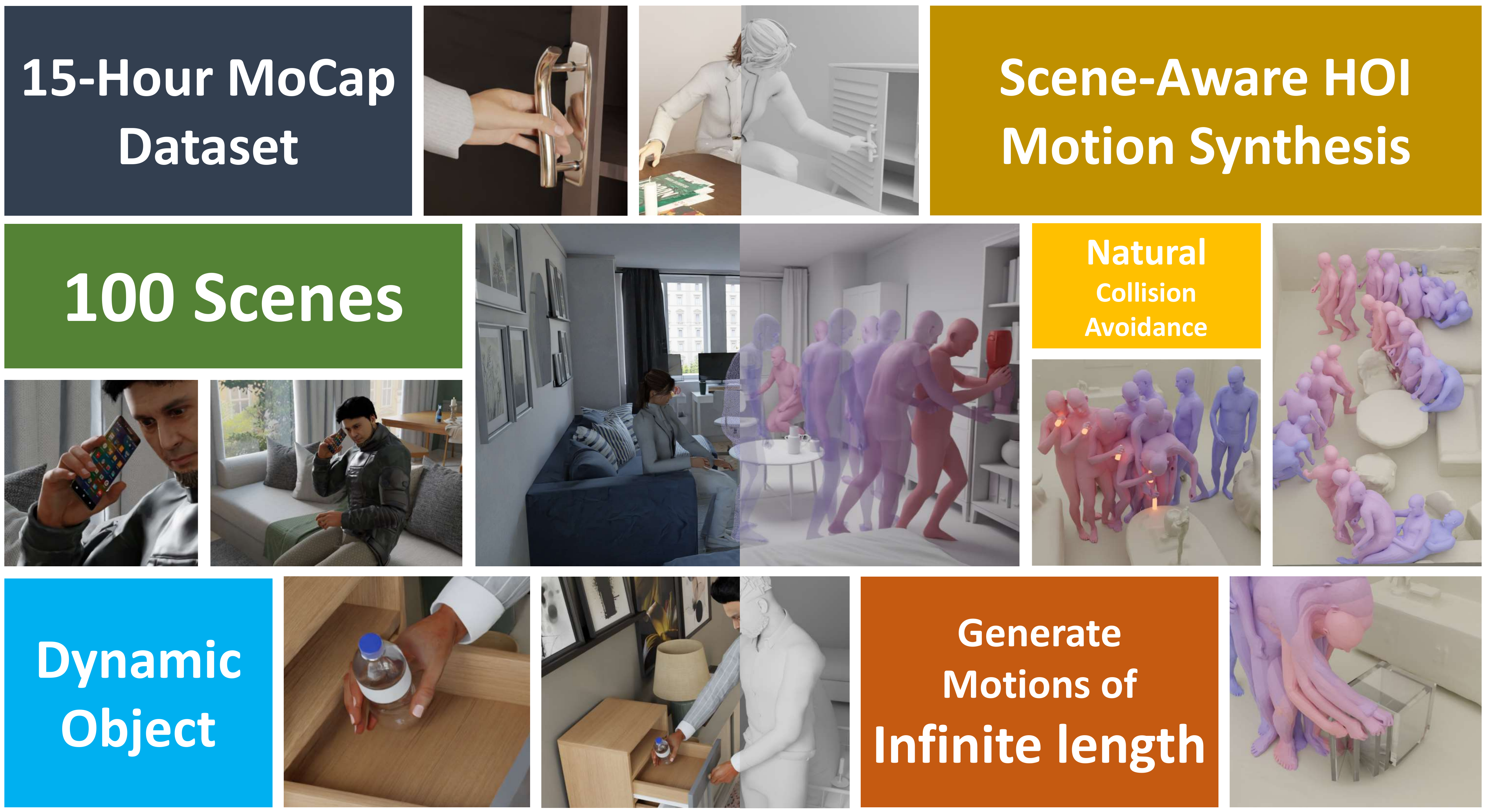} 
        \captionof{figure}{\textbf{Overview of \dataset dataset and our \acf{hsi} framework}. We introduce the most extensive motion-captured \ac{hsi} dataset, featuring diverse \acp{hsi} precisely captured in 100 scene configurations. Benefiting from \dataset, we propose a novel method for generation of \acp{hsi} with arbitrary length, surpassing all baselines and exhibiting superb zero-shot generalizability.}
        \label{fig:teaser}
    \end{center}%
}]

\begin{abstract}
\vspace{-6pt}
Confronting the challenges of data scarcity and advanced motion synthesis in \ac{hsi} modeling, we introduce the \dataset (\datasetfull) dataset alongside a novel \ac{hsi} motion synthesis method. \dataset stands as the most comprehensive motion-captured \ac{hsi} dataset currently available, encompassing over 15 hours of human interactions across 100 indoor scenes. It intricately captures whole-body human motions and part-level object dynamics, focusing on the realism of contact. This dataset is further scaled up by transforming physical environments into exact virtual models and applying extensive augmentations to appearance and motion for both humans and objects while maintaining interaction fidelity. Utilizing \dataset, we devise a diffusion-based autoregressive model that efficiently generates \acf{hsi} sequences of any length, taking into account both scene context and intended actions. In experiments, our approach shows remarkable zero-shot generalizability on a range of 3D scene datasets (\eg, PROX, Replica, ScanNet, ScanNet++), producing motions that closely mimic original motion-captured sequences, as confirmed by quantitative experiments and human studies.\vspace{-18pt}
\end{abstract}

\section{Introduction}

The intricate interplay between humans and their environment is a focal point in \acf{hsi}~\cite{gibson1950perception}, spanning diverse facets from object-level interaction~\cite{bhatnagar2022behave,lee2023locomotion} to scene-level planning and interaction~\cite{araujo2023circle,guzov23ireplica,hassan2019resolving,hassan2021populating}. While significant strides have been made, the field is notably hindered by a scarcity of high-quality datasets. Early datasets like PiGraphs~\cite{savva2016pigraphs} and PROX~\cite{hassan2019resolving} initiated the exploration but are constrained by scalability and data quality. \mocap datasets~\cite{mandery2015kit,guzov2021human} prioritize high-quality human motion capture using sophisticated equipment like \vicon. However, they often lack in capturing diverse and immersive \acp{hsi}. Scalable datasets recorded via RGBD videos offer broader utility but are impeded by lower quality in human pose and object tracking. The advent of synthetic datasets~\cite{wang2022humanise,araujo2023circle,black2023bedlam,cao2020long} provides cost efficiency and adaptability but fails to encapsulate the full spectrum of realistic \acp{hsi}, particularly in capturing dynamic 3D contacts and object tracking.

To address these challenges, this work first introduces the \dataset (\datasetfull) dataset. \dataset emerges as the most extensive motion-captured \ac{hsi} dataset, \textbf{encompassing over 15 hours of diverse human interactions across 100 indoor scenes}. It captures whole-body human motions and part-level object dynamics with an emphasis on the realism of contact. This dataset is further enhanced by digitally replicating physical environments into accurate virtual models. Extensive augmentations in appearance and motion are applied to both humans and objects, ensuring high fidelity in interaction.

Next, we devise a computational model tackling the above challenges by taking both scene and action as conditions. Specifically, our model employs an autoregressive conditional diffusion with \textbf{scene} and \textbf{action} embeddings as conditional input, capable of generating motions of arbitrary length. To integrate \textbf{scene} context, we develop an efficient local scene perceiver by querying the global scene occupancy on a localized basis, which demonstrates robust proficiency in 3D-aware collision avoidance while navigating cluttered scenes. To incorporate frame-wise \textbf{action} labels as conditions, we integrate temporal features into action segments, empowering the model to accept instructions anytime while adhering to the given action labels. This dual integration of scene and action conditions enhances the controllability of our method, providing a nuanced interface for synthesizing plausible long-term motions in 3D scenes.

We conducted a comprehensive cross-evaluation of both the \dataset dataset and our motion synthesis method. Comparing \dataset with existing ones, we demonstrate that \dataset markedly improves the performance of current state-of-the-art approaches. Moreover, our method, evaluated both qualitatively and quantitatively, exceeds existing motion synthesis methods in terms of quality and zero-shot generalizability on unseen 3D scenes, closely approximating the quality of original motion-captured data. Beyond motion synthesis, \dataset has been benchmarked for human pose and contact estimation tasks, demonstrating its versatility and establishing it as a valuable asset for a broad range of future research endeavors.

Summarized in \cref{fig:teaser}, our work significantly advances \ac{hsi} modeling. Our contributions are threefold: (i) The introduction of \dataset, an extensive \mocap \ac{hsi} dataset capturing a wide array of human behaviors across 100 indoor scenes, noted for its diversity, quality, and scalability. (ii) The development of a diffusion-based autoregressive method for the generation of \acp{hsi}, adaptable to any length and conditioned on 3D scenes and action labels. (iii) Through extensive experimentation, we demonstrate the robustness of \dataset and our proposed methods, capable of generating motions that rival \mocap quality, outperforming existing baselines, and exhibiting exceptional zero-shot generalizability in novel environments.

\section{Related Work}

\begin{table*}[ht!]
    \centering
    \small
    \setlength{\tabcolsep}{4pt}
    \caption{\textbf{Comparison of \dataset with existing \ac{hsi} datasets.} \dataset differs by providing a diverse collection of \acp{hsi}, encompassing over 15 hours of interaction across 100 indoor scenes, along with photorealistic RGBD renderings in both multi-view and ego-view.}
    \label{tab:datasets}
    \begin{tabular}{lcccccccccc}
        \toprule
        \multirow{2}{*}{Datasets} & \multirow{2}{*}{Hours} & \multirow{2}{*}{MoCap}  & Human & Dynamic & No. of & Contact & \multirow{2}{*}{RGBD} & \multirow{2}{*}{Segmentation} & Multi- & Ego- \\
        &  & & Representation & Object & Scenes & Annotations & &  & view & view \\
        \midrule
        GTA\_IM~\cite{cao2020long}          & 9.3  &        & skeleton  &        & 10   &           & \cmark    &       & \cmark    &           \\
        PiGraphs~\cite{savva2016pigraphs}   & 2.0  &        & skeleton  &        & 30   &           & \cmark    &       &           &           \\  
        PROX~\cite{hassan2019resolving}     & 0.9  &        & SMPL-X    &        & 12   & \cmark    & \cmark    & \cmark&           &           \\  
        GRAB~\cite{taheri2020grab}          & 3.8  & \cmark & SMPL-X    & \cmark & -    & \cmark    &           &       &           &           \\  
        SAMP~\cite{hassan2021stochastic}    & 1.7  & \cmark & SMPL-X    &        & -    &           &           &       & \cmark    &           \\  
        RICH~\cite{huang2022capturing}      & 0.8  &        & SMPL-X    &        & 5    & \cmark    & \cmark    &       & \cmark    &           \\  
        BEHAVE~\cite{bhatnagar2022behave}   & 4.2  &        & SMPL      & \cmark & -    & \cmark    & \cmark    & \cmark& \cmark    &           \\  
        CHAIRS~\cite{jiang2023full}         & 17.3 & \cmark & SMPL-X    & \cmark & -    & \cmark    &  \cmark   &       & \cmark    &           \\
        COUCH~\cite{zhang2022couch}         & 3.0  & \cmark & SMPL      & \cmark & -    & \cmark    & \cmark    & \cmark& \cmark    &           \\
        iReplica~\cite{guzov23ireplica}     & 0.8  & \cmark & SMPL      & \cmark & 7    & \cmark    & \cmark    &       & \cmark    & \cmark    \\  
        CIRCLE~\cite{araujo2023circle}      & 10.0 & \cmark & SMPL-X    &        & 9    &           &           &       &           & \cmark    \\  
        \midrule
        \rowcolor{LightGray}
        \dataset                            & 15.0 & \cmark & SMPL-X    & \cmark & 100  & \cmark    & \cmark    & \cmark& \cmark    & \cmark \\ 
        \bottomrule
    \end{tabular}%
\end{table*}

\paragraph{\ac{hsi} Datasets}

Capturing human motions in 3D scenes is pivotal, with an emphasis on the quality and scale of human interactions. Early work focused on capturing coarse 3D human motions using 2D keypoints~\cite{monszpart2019imapper} or RGBD videos~\cite{savva2016pigraphs}. To improve quality and granularity, datasets like PROX~\cite{hassan2019resolving} use scene scans as constraints to estimate SMPL-X parameters~\cite{pavlakos2019expressive} from RGBD videos. However, these image-based motion capture methods often result in noisy 3D poses.

Recent efforts have incorporated more sophisticated systems like IMU or optical \mocap (\eg, \vicon)~\cite{mandery2015kit,guzov2021human,zhang2022couch,hassan2021stochastic,jiang2023full,guzov23ireplica}, providing higher quality capture but limited in scalability. These are typically constrained to static scenes~\cite{wang2022humanise,hassan2021stochastic,guzov23ireplica} or single objects~\cite{zhang2022couch,bhatnagar2022behave,jiang2023full}, not fully representing complex real-world \acp{hsi} such as navigating cluttered spaces or managing concurrent actions.

Synthetic datasets~\cite{wang2022humanise,araujo2023circle,cao2020long} have attempted to fill this gap. Notable examples like BEDLAM~\cite{black2023bedlam} and CIRCLE~\cite{araujo2023circle} have been acknowledged for their cost efficiency and adaptability. These datasets integrate human motion data into synthetic scenes but fail to fully capture the range of realistic 3D \acp{hsi}, particularly in terms of dynamic object poses within their simulated environments.

Addressing these shortcomings, our work achieves a unique balance of quality and scalability. We replicate synthetic 3D environments in an optical motion capture setting, facilitating both accurate capture of humans and objects in complex \acp{hsi} and providing photorealistic renderings. This approach not only enhances the fidelity of the captured interactions but also extends the range of scenarios and environments that can be realistically simulated.

\paragraph{\ac{hsi} Generation}

\ac{hsi} generation involves single-frame human body~\cite{li2019putting,zhang2020generating,zhangsiwei2020generating} and temporal motion sequences~\cite{wang2021synthesizing,wang2021scene,hassan2021stochastic,wang2022towards,huang2023sceneDiffuser,araujo2023circle,pan2023synthesizing,mir2023generating,xu2023interdiff,li2023object}, utilizing models like \ac{cvae}~\cite{sohn2015learning} and diffusion models~\cite{sohl2015deep,song2019generative,ho2020denoising}. Recent advancements focus on generating arbitrary-length human motions through autoregressive methods~\cite{corona2020context,cao2020long,hassan2021stochastic,taheri2021Goal,mao2022contactMotionForecasting,zhang2023generating} and anchor frame generation~\cite{wang2021synthesizing,pi2023hierarchical}. Additionally, enhancing generation controllability has involved semantic guidance, such as action labels~\cite{zhao2022compositional} and language descriptions~\cite{wang2022humanise,xiao2023unified}.

In comparison, our work contributes a conditional generative model with an autoregressive mechanism to generate \textbf{arbitrary-length} motions, combining diffusion model capabilities with improved \textbf{controllability} in \ac{hsi} generation.

\begin{figure*}[t!]
    \centering
    \includegraphics[width=\linewidth]{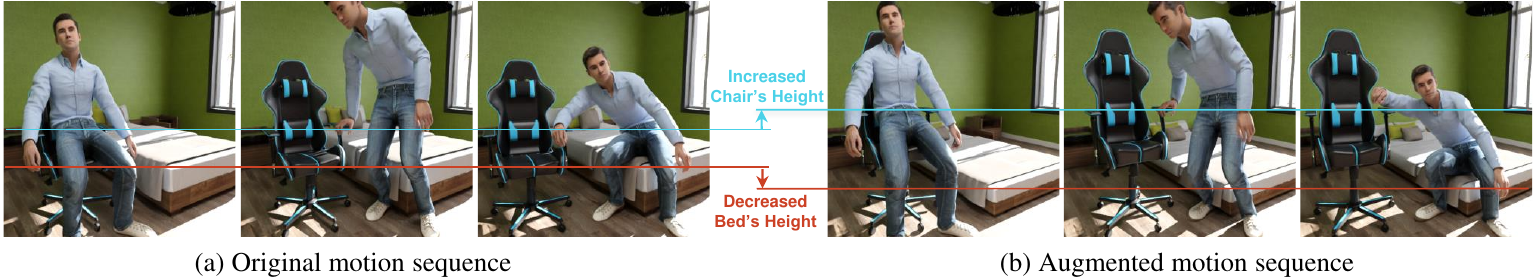}
    \caption{\textbf{Data augmentation for motion generation.} This example highlights how human motion is adjusted to accommodate variations in object sizes. Specifically, the chair's height is increased, and the bed's height is decreased, each by $15$cm. Our augmentation method proficiently modifies human motion to maintain consistent interactions despite these changes in object dimensions.}
    \label{fig:nsm}
\end{figure*}%

\section{\texorpdfstring{\dataset}{} Dataset}

This section introduces \dataset, the most comprehensive \mocap dataset dedicated to 3D \acp{hsi} thus far. \dataset offers not only accurate 3D ground truths but also photorealistic renderings accompanied by various 2D ground truths, suitable for various perceptual tasks in \ac{hsi}. This section details the dataset's statistics, data capture process, post-processing method, and our augmentation pipeline.

\subsection{Dataset Statistics}

\dataset encompasses 15 hours of high-quality motion-captured data, featuring complex \acp{hsi} within 3D scenes, where humans interact with clustered environments and dynamic objects. Captured at a rate of 30 Hz using the state-of-the-art \vicon \mocap system, the dataset comprises a total of 1.6 million frames.
The \ac{hsi} interactions in \dataset include 20 different types of common objects, ensuring a minimum of 5 distinct instances per type. The object categories encompass a range from static items like sofas and beds to dynamic objects such as bottles, and even articulated items including laptops and cabinets.
\dataset incorporates performances from 7 participants (4 male and 3 female), who enacted various actions across 100 indoor scenes. These scenes span a variety of settings, such as dining rooms, living rooms, bedrooms, and kitchens, among others. For a comprehensive comparison of the \dataset dataset with existing \ac{hsi} datasets, please refer to \cref{tab:datasets}.

\subsection{Scene-aware Motion Capture}

Aiming to capture realistic and diverse \acf{hsi} within 3D scenes, our approach emphasizes both data quality and diversity. We initiate this process by replicating 3D scenes and objects sourced from the 3D-FRONT~\cite{fu20213dfront} dataset and BlenderKit~\cite{blenderkit} within the physical environment housing our \mocap devices. 
To ensure the naturalness of human interactions during motion capture, we meticulously create real-world placeholders that correspond to the affordances of the objects in the synthetic environment. All movable objects are tagged with markers compatible with the \vicon system, enabling precise tracking of their poses. Actors undergo training to familiarize themselves with interacting with these placeholders. During the capturing sessions, actors are prompted to perform actions randomly selected from a pre-defined pool, ensuring a variety of interactions.

Post-capture, the human poses are converted into the SMPL-X format~\cite{pavlakos2019expressive}, employing a vertex-to-vertex optimization technique. This method is instrumental in calculating vertex-to-vertex distances between the human meshes and object meshes, facilitating accurate per-vertex contact annotations. We utilize Blender~\cite{blender} to render multi-view photorealistic RGBD videos, segmentation masks, and ego-centric videos. To further diversify the renderings, we incorporate over 200 digital human models from Character Creator 4~\cite{CharacterCreator4}, ensuring that objects strategically placed in scene backgrounds enhance the scene's realism without impeding human movement. For a detailed exposition of our capture and processing pipeline, refer to \cref{sec:supp:capture}.

\subsection{\mocap Data Augmentation}

Our data augmentation pipeline is designed to adapt human motions to changes in 3D scene objects, ensuring physical plausibility and accuracy in \ac{hsi}, following~\cite{starke2019neural}. This process is vital in complex scenarios with concurrent or successive interactions; see \cref{fig:nsm}. The pipeline consists of three main steps for integrating altered human motions into diverse 3D settings.

\paragraph{Calculate Target Joint}

We identify contact points between human joints and object meshes, and locate corresponding points on transformed or replaced objects. This step crucially adjusts the target joint's position to maintain the original interaction's contact relationship, ensuring realistic human-object interactions despite changes in object dimensions or positions.

\paragraph{Refine Trajectory}

To smooth out abrupt trajectory changes from the first step or \ac{ik} computations, we apply temporal smoothing to joint offsets, iteratively adjusting weights in adjacent frames. This refinement is critical for maintaining seamless motion, particularly in scenarios with multiple object interactions. Further details and theoretical background are discussed in \cref{sec:supp:nsm_detail}.

\paragraph{Recompute Motion with \ac{ik}}

In the final step, we recompute human motion using the smoothed trajectories with an enhanced CCD-based~\cite{kenwright2012inverse} \ac{ik} solver. This solver applies clipping and regularizations to bone movements, ensuring natural motion fluidity. Bones further from the root joint have increased rotational limits, reducing jitteriness and enhancing motion realism. For a complete description of these methods, refer to \cref{sec:supp:nsm_detail}.

\begin{figure*}[t!]
    \centering
    \includegraphics[width=\linewidth]{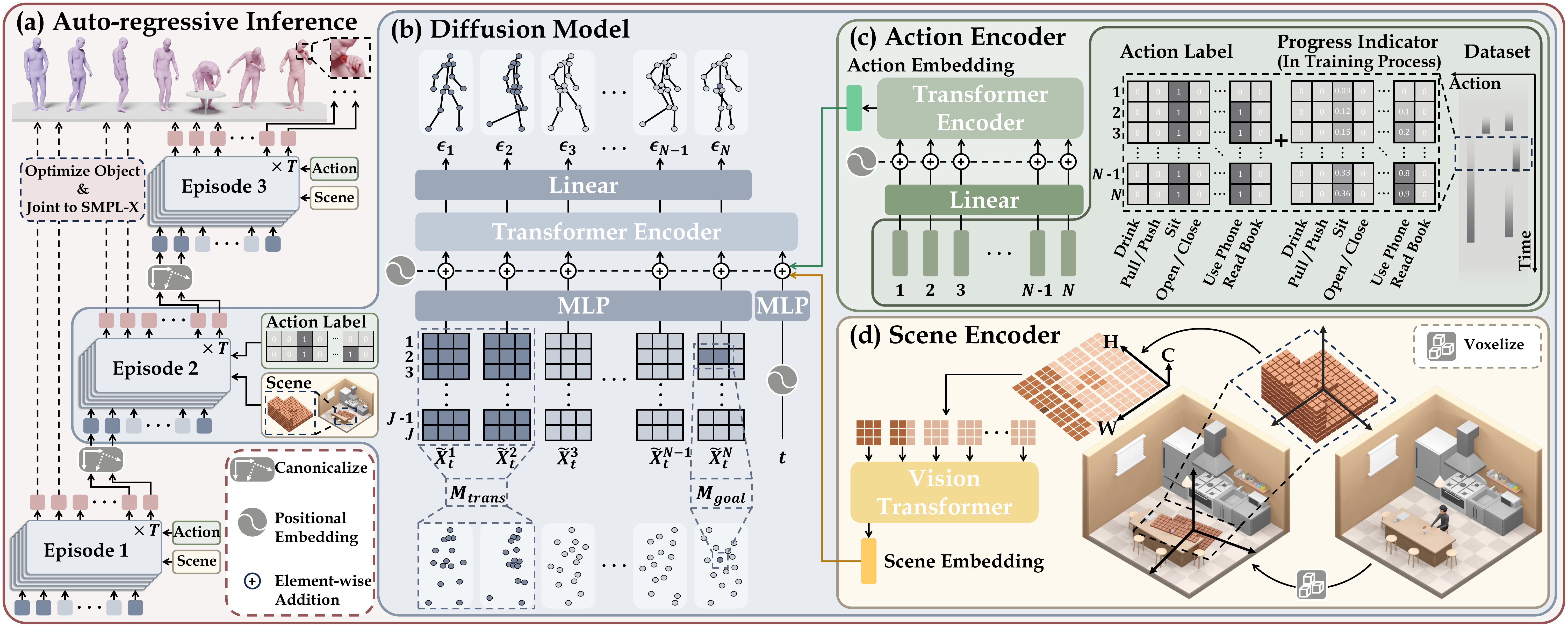}
    \caption{\textbf{Model architecture.} (a) Our model employs an autoregressive diffusion sampling approach to generate arbitrary long-sequence motions. (b) Within each episode, we synthesize motion using DDPM integrated with a transformer architecture, taking the human joint locations as input. (c)(d) Action and scene conditions are encoded and forwarded to the first token, guiding the motion synthesis process.}
    \label{fig:model}
\end{figure*}

\section{Method}

Utilizing the comprehensive \dataset dataset, we develop an autoregressive motion diffusion model. This model generates \acp{hsi} that are not only physically plausible in 3D scenes but also highly \textbf{controllable} through frame-wise action labels, capable of producing sequences of \textbf{arbitrary} length in \textbf{real-time}. 

\subsection{Problem Formulation and Notations}

Given a 3D scene $\mathcal{S}$, a goal location $\mathcal{G}$, and action labels $\mathcal{A}$, our objective is to synthesize a human motion sequence $\{\mathcal{H}_i\}_{i=1}^{L}$ of arbitrary length $L$. When interacting with dynamic objects $\mathbf{P}$, we also estimate the corresponding object pose sequence $\{\mathcal{O}_i\}_{i=1}^{L}$.

\paragraph{Human}

Human motion is represented as a sequence of parameterized human meshes $\{\mathcal{H}_i\}$ using the SMPL-X model~\cite{pavlakos2019expressive}. The motion is initially generated as body joints locations $\{X^i\}_{i=1}^{L}$, where $X^i \in \mathbb{R}^{J\times 3}$ represents $J=24$ selected joints. These are fitted into the SMPL-X pose parameters $\theta$, global orientation $\phi$, hand poses $h$, and root translation $r$, resulting in the posed human mesh $\mathcal{H}\in\mathbb{R}^{10475 \times 3}$.

\paragraph{Conditions}

We formalize three types of conditions in our motion synthesis: 3D scene, goal location, and action labels. The 3D scene is represented by a voxel grid $\mathcal{S}\in\{0, 1\}^{N_x \times N_y \times N_z}$, with $1$ indicating reachable locations. Goal locations are 2D positions $\mathcal{G} \in \mathbb{R}^{2}$ for navigation, or 3D $\mathbb{R}^{3}$ for joint-specific control. Action labels are multi-hot vectors $\mathcal{A} \in \{0, 1\}^{L \times N_A}$, indicating distinct actions.

\paragraph{Object}

When dynamic objects are involved, the object is represented by its point cloud $\mathbf{P}$ in canonical coordinates and its global rotation $R$ and translation $T$. The dynamic object sequence $\{O_i\}_{i=1}^{L}$ is then represented by sequences of rotations and translations $\{R_i, T_i\}_{i=1}^{L}$.

\subsection{Autoregressive Motion Diffusion}

Our model architecture is illustrated in \cref{fig:model}. Our goal is to generate human motions that are not only physically plausible in 3D scenes but also highly controllable by frame-wise action labels, achieving arbitrary length in real time. We employ an autoregressive diffusion strategy where a long motion sequence is progressively generated by \textit{episodes}, each defined as a motion segment of $L_{epi}$ frames. Based on the approach by \citet{shafir2023human}, successive episodes are generated by extending from the final $k$ frames of the prior episode. For each new episode, the first $k$ frames are set based on the previous episode's last $k$ frames, with the noise on these transition frames zeroed out using a mask $\mathbf{M}_{trans}$. Our model aims to inpaint the remainder of each episode by filling in the unmasked frames.

To ensure precise control over character navigation and detailed interactions in each episode, we segment the overall goal $\mathcal{G}$ into discrete subgoals, represented as ${\{\mathcal{G}_i\}_{i=1}^{N_{epi}}}$, where $N_{epi}$ denotes the number of episodes. For navigation, each subgoal $\mathcal{G}_i\in\mathbb{R}^2$ dictates the desired $xy$-coordinates of the character's pelvis at an episode's conclusion. Mirroring the masking approach used in $\mathbf{M}_{trans}$, we align the pelvis's $xy$-coordinate in the episode's final frame to the respective subgoal, simultaneously masking the corresponding diffusion noise. As the $z$-coordinate is unspecified, the model is trained to infer the appropriate pelvis height based on the scene setup, such as making the character sit when the subgoal indicates a chair's location. This principle also governs fine-grained interactions, like grasping or pushing, where the subgoal $\mathcal{G}_i\in\mathbb{R}^3$ is set to the precise 3D location, aligning the relevant hand joint to $\mathcal{G}_i$ and masking joint noise accordingly. This specific masking on the subgoals is denoted as $\mathbf{M}_{goal}$.

We devise a conditional diffusion model for generating motions within each episode. This process involves sampling from a Markov noising process $\{X_{t}\}_{t=0}^{T}$. Starting with the original human joint data $X_0$ drawn from the data distribution, Gaussian noise is added to the components of $X_0$ not masked by $\mathbf{M}=\mathbf{M}_{trans} \cup \mathbf{M}_{goal}$. The unmasked components, represented as $(1-\mathbf{M}) \odot X_t$ or $\tilde{X}_{t}$ (where $\odot$ is the Hadamard product), undergo a forward noising process
\begin{equation}
    q(\tilde{X}_{t} | \tilde{X}_{t-1}) = \mathcal{N}(\tilde{X}_{t};\sqrt{\alpha_{t}}\tilde{X}_{t-1},(1-\alpha_{t})I),
\end{equation}
with $\alpha_{t} \in (0,1)$ denoting hyper-parameters related to the variance schedule.

Motion data generation within our model employs a reversed diffusion process to gradually denoise $\tilde{X}_T$. Consistent with established diffusion model training methodologies, noise $\epsilon_t$ is applied to obtain $\tilde{X}_{t}$, and a neural network $\epsilon_\theta(\tilde{X}_{t}, t, \mathcal{S}, \mathcal{A})$ is constructed to approximate this noise. The learning objective for $\epsilon_\theta$ follows a simple objective~\cite{ho2020denoising}
\begin{equation}
    \mathcal{L}=E_{\tilde{X}_0\sim q(\tilde{X}_0|\mathcal{C}),t\sim[1,T]}\norm{\epsilon - \epsilon_\theta(\tilde{X}_t, t, \mathcal{S}, \mathcal{A})}_{2}^2.
\end{equation}

We adopt the Transformer model architecture~\cite{tevet2022human}, wherein the first token encodes information about the diffusion step, scene, and action, and subsequent tokens represent the noisy joint locations for each frame in the current episode. Throughout the sampling process, the model predicts the noise applied to each joint element. Once this sampling phase concludes, the joint locations are translated into SMPL-X parameters via a lightweight MLP. This translation is further refined through an optimization process, ensuring accurate alignment with the human joint data.

Upon generating the human motion sequence $\{\mathcal{H}_i\}_{i=0}^L$, we optimize the trajectory of the interacting object $\{\mathcal{O}_i\}_{i=0}^L$ to ensure natural \acp{hoi}. To enhance the realism of the interaction, we further fine-tune the object's pose in each frame to minimize the variance in distance between the object and the interacting hand~\cite{ghosh2023imos}.

\subsection{Local Scene Perceiver}

As illustrated in \cref{fig:model}(d), the local scene perceiver is essential for embedding the local scene context, serving as a condition for motion generation. This component analyzes the scene using a local occupancy grid centered around the subgoal location for the current episode. Starting with the global occupancy grid $\mathcal{S}$ of the scene, where each cell's boolean value indicates reachability (1 for reachable, 0 otherwise), we focus on the $i$-th episode's subgoal $\mathcal{G}_i=(x,y,z)$ or $(x,y)$. A local occupancy grid is constructed around $(x,y)$, extending vertically from 0 to 1.8m. The grid's orientation aligns with the yaw of the agent's pelvis at the episode's start, and cell values are derived by querying the global occupancy grid.

The voxel grid is encoded using a Vision Transformer (ViT)~\cite{dosovitskiy2020vit}. We prepare the tokens by dividing the local occupancy grid into patches along the $xy$-plane, considering the $z$-axis as feature channels. These patches are then fed into the ViT model. The resulting scene embedding from this process is utilized as the condition for the diffusion model.

Discretizing the scene into a grid format is a necessary trade-off to boost training efficiency and practicality in our \ac{hsi} method. Although directly generating the local occupancy grid from the scene mesh in real-time is technically feasible, it substantially prolongs training time. For instance, employing the \textit{checksign} function from Kaolin results in a training process that is approximately 300 times slower, rendering it impractical. Despite this simplification, our empirical results demonstrate that the quality of motion generation is not significantly impacted by this approximation.

\subsection{Frame-wise Action Embedding}

Our method distinguishes itself from prior approaches by incorporating frame-wise action labels into the long-term motion synthesis process, rather than generating a complete motion sequence from a singular action description. In our framework, a particular action can span multiple episodes, necessitating the model's capability to comprehend the evolution and progression of an action over time.

To enhance our model's understanding of action progression, we incorporate a progress indicator $\mathcal{A}_{ind}\in \mathbb{R}^{L_{epi} \times N_A}$ into the frame-wise action labels, as depicted in \cref{fig:model}(c). This indicator is realized by appending a real number $n\in[0,1]$ to the original action labels, representing the action's advancement from start to finish. As a result, action labels take on values in ${0 \cup [1, 2]}$ post-addition. For instance, during a drinking action from frame $i$ to $j$, we modify the $(0, 1)$ label by adding a value that linearly progresses from 0 to 1 across this interval. Thus, at the onset of drinking (frame $i$), the label is augmented to 1, gradually increasing to 2 by frame $j$, the action's conclusion. This nuanced labeling enables our model to seamlessly handle actions that span multiple episodes, significantly enhancing the realism and fluidity of the synthesized motion sequences.

The final action embedding is obtained by processing the progress-augmented action label $\mathcal{A} \in \mathbb{R}^{L_{epi} \times N_A}$ through a Transformer encoder. Each frame's action label $\mathcal{A}_i \in \mathbb{R}^{N_A}$ is treated as an individual token in the Transformer's input. The feature output from the last token is then passed through an MLP to generate the final action embedding.

\section{Experiments}

This section presents our evaluation of both \dataset and our proposed motion synthesis method, focusing on action-conditioned \ac{hsi} generation. Additionally, we demonstrate how \dataset contributes to advancements in state-of-the-art motion synthesis methods.

\subsection{Experiment Settings}

Our experimental evaluation of \ac{hsi} generation quality is conducted under two distinct settings: \textit{static} and \textit{dynamic}. The \textit{static} setting assesses synthesized motions in environments without dynamic interactable objects, concentrating on locomotion and interactions with static objects. Conversely, the \textit{dynamic} setting evaluates motion synthesis involving interactions with dynamic objects. In both scenarios, we compare the performance of methods trained on \dataset with those trained on existing datasets~\cite{zhang2020generating,taheri2020grab}, offering a thorough insight into both the model's efficacy and the dataset's impact.

\subsection{Baselines and Ablations}

\paragraph{Baselines--\textit{static} setting}

We compare \dataset with PROX~\cite{zhang2020generating}, a dataset featuring human activities in indoor scenes. To ensure a fair comparison, we retain only the locomotion and scene interaction of static objects in \dataset, such as sitting and lying down. Baseline methods for this setting include cVAE~\cite{wang2021synthesizing}, SceneDiff~\cite{huang2023sceneDiffuser}, and GMD~\cite{karunratanakul2023GMD}.

\paragraph{Baselines--\textit{dynamic} setting}

We compare \dataset with GRAB~\cite{taheri2020grab}, known for capturing full-body grasping actions with human and object pose sequences. Here, the focus is on motions of interaction with dynamic objects, like drinking water and making phone calls, present in both datasets. We compare our method against IMoS~\cite{ghosh2023imos} and GOAL~\cite{taheri2021Goal}, reproduced using their original implementations.

\paragraph{Ablations}

In our ablative studies, we examine the impact of disabling the action progress indicator $\mathcal{A}_{ind}$ in our model. Additionally, to assess the significance of our data augmentation technique, we perform experiments using a non-augmented version of \dataset. For reference, our standard experiments employ the augmented \dataset, where each object is transformed into two different variations.

Our evaluation encompasses 10 unseen indoor scenes sourced from PROX~\cite{hassan2019resolving}, Replica~\cite{straub2019replica}, Scannet~\cite{dai2017scannet}, and Scannet++~\cite{yeshwanth2023scannet++}. These scenes are adapted to the requirements of different methods, with modifications including conversion to point cloud format, voxelization, or maintaining their original mesh format. To evaluate the diversity of the synthesized motions, each method is tasked with generating five unique variations for each trajectory.

Furthermore, we conduct a qualitative comparison of our method with other recent approaches, such as SAMP~\cite{hassan2021stochastic}, DIMOS~\cite{zhao2023synthesizing}, LAMA~\cite{lee2023locomotion}, and \citet{wang2022towards}, based on the feasibility of reproducing these methods. Detailed findings from this comparison are discussed in \cref{sec:supp:human-study}.

\begin{figure*}[t!]
    \centering
    \includegraphics[width=\linewidth]{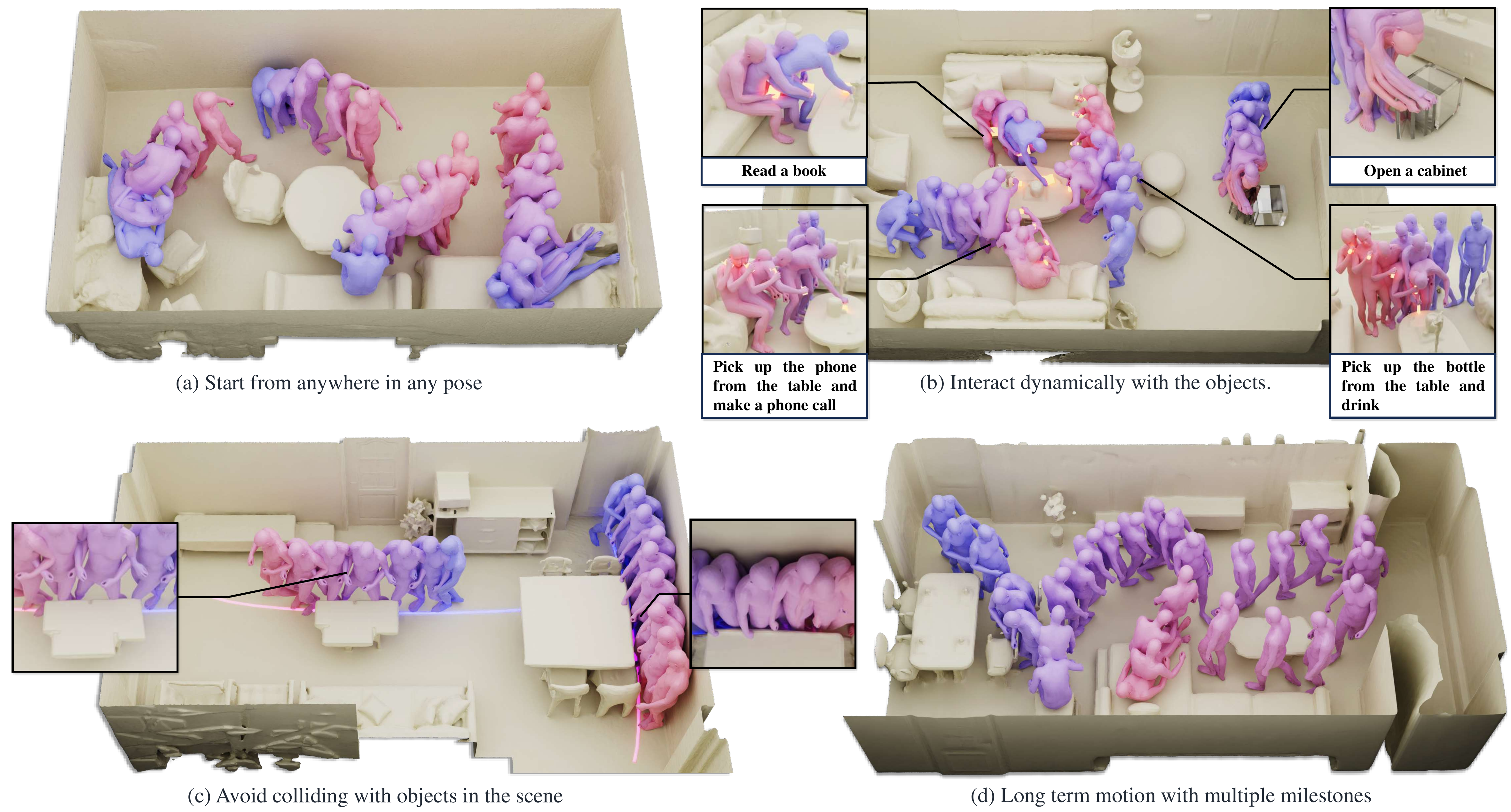}
    \caption{\textbf{Visualization of motion generation.} Leveraging local scene context and action instructions as conditions, our method demonstrates its proficiency in (a) initiating motion given the surrounding environment, (b) dynamically interacting with objects, (c) avoiding collisions during motion progression, and (d) robustly synthesizing long-term motion. The depicted scenes are selected from PROX, Replica, and FRONT3D-test datasets, none of which were included in the training phase. For qualitative results, please refer to the \textit{Supplementary Video}.}
    \label{fig:results}
\end{figure*}

\subsection{Evaluation Metrics}

In the \textit{static} setting, we employ \textit{Contact} and \textit{Penetration} metrics, as recommended by \citet{zhao2023synthesizing}, to evaluate foot slide and object penetration issues in synthesized motions. These metrics measure the degree to which the synthesized motions conform to the specified scene. For the \textit{dynamic} setting, we utilize \textit{FID} and \textit{Diversity} metrics, commonly used in language and action-guided motion generation tasks~\cite{tevet2022human,ghosh2023imos}. These metrics measure the quality and diversity of \ac{hoi} motion generation involving various small objects.

Additionally, we introduce a novel MoCap-differentiating human study for evaluation. Participants are presented with five sequences, one of which is motion-captured, and are asked to identify the MoCap sequence. The likelihood of correctly identifying the MoCap sequence serves as an indicator of the synthesized motion's realism. We quantify this aspect through the Success Rate of Discrimination (SucRateDis), reflecting the percentage of participants who accurately identify the MoCap sequence.

\subsection{Results and Analysis}

\cref{fig:results} showcases our method's qualitative strengths. It adeptly manages complex scene configurations, including initiating context-aware motion, avoiding collisions during movement, and generating extended motions, especially in \ac{hoi} scenarios involving dynamic object interaction.

In the \textit{static} setting (\cref{tab:results:nondynamic}), our method, trained on \dataset, surpasses baselines across most metrics. Notably, disabling data augmentation leads to increased penetration, suggesting the efficacy of augmented data in producing physically plausible motions. Compared to models trained on PROX, ours shows significant improvements, highlighting \dataset as a high-quality resource for \ac{hsi} research.

\begin{table}[ht!]
    \centering
    \small
    \setlength{\tabcolsep}{4pt}
    \caption{\textbf{Evaluation of locomotion and scene-level interaction}. We compare performances on \textcolor{Pink}{\dataset} and \textcolor{Cyan}{PROX}~\cite{hassan2019resolving}.}
    \label{tab:results:nondynamic}
    \resizebox{\linewidth}{!}{
        \begin{tabular}{lcccc}
            \toprule
            Method                                                       & Cont.$\uparrow$  & $\text{Pene}_{mean}\downarrow$ & $\text{Pene}_{max}\downarrow$ & Dis. suc.$\downarrow$ \\
            \midrule
            \rowcolor{MistyRose} \citet{wang2021synthesizing}            & 0.969            & 1.935                          & 14.33                         & 0.581  \\
            \rowcolor{MistyRose} SceneDiff~\cite{huang2023sceneDiffuser} & 0.912            & \textbf{1.691}                 & 17.48                         & 0.645  \\
            \rowcolor{MistyRose} GMD~\cite{karunratanakul2023GMD}        & 0.931            & 2.867                          & 21.30                         & 0.871  \\
            \rowcolor{MistyRose} Ours                                    & \textbf{0.992}   & 1.820                          & \textbf{11.74}                & \textbf{0.258} \\  
            \rowcolor{MistyRose} Ours w/o aug.                             & 0.991            & 2.010                          & 15.52                         & - \\  
            \midrule
            \rowcolor{LightCyan1} \citet{wang2021synthesizing}              & 0.688            & 4.935                          & 34.10                         & 0.903 \\
            \rowcolor{LightCyan1} SceneDiff~\cite{huang2023sceneDiffuser}   & 0.712            & 3.267                          & 27.48                         & 0.935 \\
            \rowcolor{LightCyan1} GMD~\cite{karunratanakul2023GMD}          & 0.702            & 4.867                          & 38.30                         & 0.968 \\
            \rowcolor{LightCyan1} Ours                                      & 0.723            & 4.820                          & 31.74                         & 0.903 \\
            \bottomrule
        \end{tabular}
    }
\end{table}

\cref{tab:results:dynamic} illustrates results in the \textit{dynamic} setting, where our approach excels in 3D \ac{hoi} generation. High penetration rates with GRAB-trained methods indicate its limitations in scene-adherent \ac{hoi} motions, while \dataset captures more detailed interactions. The absence of the progress indicator $\mathcal{A}_{ind}$ leads to method failure, as evidenced by the ablation study.

\begin{table}[ht!]
    \centering
    \small
    \setlength{\tabcolsep}{4pt}
    \caption{\textbf{Evaluation of object-level interaction.} We compare performances on \textcolor{Pink}{\dataset} and \textcolor{Cyan}{GRAB}~\cite{taheri2020grab}. The definition of ``Real'' follows the one defined in \citet{tevet2022human}}
    \label{tab:results:dynamic}
    \resizebox{\linewidth}{!}{
        \begin{tabular}{lcccc}
            \toprule
            Method                                             & FID$\downarrow$ & Diversity$\rightarrow$ & $\text{Pene}_{scene}\downarrow$  & Dis. suc.$\downarrow$ \\
            \midrule
            \rowcolor{MistyRose} Real-\dataset                 & -               & 2.734                  & -                                & - \\
            \rowcolor{MistyRose} GOAL~\cite{taheri2021Goal}    & 0.512           & 2.493                  & 34.10                            & 0.801 \\
            \rowcolor{MistyRose} IMoS~\cite{ghosh2023imos}     & 0.711           & 2.667                  & 37.48                            & 0.774 \\
            \rowcolor{MistyRose} Ours                          & \textbf{0.313}  & \textbf{2.693}         & 11.74                            & \textbf{0.226}\\   
            \rowcolor{MistyRose} Ours - $\mathcal{A}_{ind}$    & 2.104           & 1.318                  & \textbf{10.62}                   & 1.000 \\ 
            \midrule
            \rowcolor{LightCyan1} Real-GRAB~\cite{taheri2020grab} & -               & 2.155                  & -                                & - \\
            \rowcolor{LightCyan1} GOAL~\cite{taheri2021Goal}      & 0.429           & 2.180                  & 44.09                            & 0.801 \\
            \rowcolor{LightCyan1} IMoS~\cite{ghosh2023imos}       & 0.410           & 2.114                  & 41.50                            & 0.774 \\
            \rowcolor{LightCyan1} Ours                            & 0.362           & 2.150                  & 34.41                            & 0.516 \\   
            \bottomrule
        \end{tabular}
    }
\end{table}

Human studies further affirm the quality of our method. Only about a quarter of participants could distinguish our synthesized motions from real MoCap data, nearly aligning with the 1/5 SucRateDis of random guessing. This suggests that our synthesized motions are nearly indistinguishable from high-quality MoCap data. Comparative evaluations with recent methods~\cite{hassan2021stochastic,zhao2023synthesizing,lee2023locomotion,wang2022towards} show our model's superiority, outperforming the second-best model by over 30\% in support rate. For more detailed results, please refer to the \textit{Supplementary Video}.

\paragraph{Real-time Control}

Our method can sample an episode of motion (1.6 seconds at 10 FPS) in 0.7 seconds on an A800 GPU. This efficiency enables uninterrupted long-term motion generation with a consistent control signal. For new control signals, to minimize the 0.7-second delay, we implement an incremental sampling strategy: initially, 2 frames are sampled immediately, followed by sampling 4 frames during their execution, increasing exponentially until 16 frames are sampled. This approach ensures a balance between real-time control and smooth motion continuity. Please refer to our \textbf{Supplementary Video} for a visual demonstration.

\subsection{Additional Image-based Tasks}

\dataset, with its photo-realistic renderings and per-vertex 3D contact annotations, is also suited for various image-based tasks. We focus on its application in 3D human mesh estimation and contact estimation.

\paragraph{3D Human Mesh Estimation}

For reconstructing 3D human body meshes from input images, we utilize the state-of-the-art method~\cite{ma20233d} as a baseline. We evaluate if including \dataset in training enhances performance on the 3DPW dataset~\cite{von2018recovering}. Following \citet{ma20233d}, we report MPJPE, PA-MPJPE, and MPVE for the estimated poses and meshes.

\paragraph{3D Contact Estimation}

This task involves predicting per-vertex 3D contact on the SMPL mesh~\cite{smpl2015} from an input image. We compare \dataset against RICH~\cite{huang2022capturing} and DAMON~\cite{tripathi2023deco}, both featuring vertex-level 3D contact labels with RGB images. Utilizing BSTRO~\cite{huang2022capturing} for RICH and DECO~\cite{tripathi2023deco} for DAMON, we measure precision, recall, F1 score, and geodesic error following the literature~\cite{huang2022capturing,tripathi2023deco}.

\paragraph{Results and Analysis}

Quantitative results in \cref{tab:mesh_estimation} reveal that integrating \dataset with 3DPW significantly improves human mesh estimation. Contact estimation outcomes, presented in \cref{tab:results:contact}, show enhanced performance with \dataset, particularly in reducing geodesic error. These results suggest that combining synthetic data from \dataset with real-world data substantially benefits image-based tasks. For detailed experimental insights, see \cref{sec:supp:image-based-tasks}.

\begin{table}[ht!]
    \centering
    \small
    \setlength{\tabcolsep}{4pt}
    \caption{\textbf{Performance of \citet{ma20233d} trained on 3DPW~\cite{von2018recovering} combined with \dataset in different ratios.}}
    \label{tab:mesh_estimation}
    \begin{tabular}{lccc}
        \toprule
        Training Data                 & MPVE\(\downarrow\) & MPJPE\(\downarrow\) & PA-MPJPE\(\downarrow\) \\
        \midrule
        3DPW~\cite{von2018recovering} & 101.3              & 88.2                & 54.4 \\
        3DPW+T (2:1)                  & 88.8               & \textbf{77.2}       & \textbf{46.4} \\
        3DPW+T (1:1)                  &  \textbf{78.5}     &  78.5               & \textbf{46.4} \\
        \bottomrule 
    \end{tabular}
\end{table}
\vspace{-10pt}

\begin{table}[ht!]
    \centering
    \small
    \setlength{\tabcolsep}{4pt}
    \caption{\textbf{Performance of \textcolor{Pink}{BSTRO}~\cite{huang2022capturing} and \textcolor{Cyan}{DECO}~\cite{tripathi2023deco} trained on \textcolor{Pink}{RICH}~\cite{huang2022capturing} and \textcolor{Cyan}{DAMON}~\cite{tripathi2023deco} combined with \dataset, respectively.}}
    \label{tab:results:contact}
    \begin{tabular}{lcccc}
        \toprule
        Training Data                                       & Prec\(\uparrow\) & Rec\(\uparrow\) & F1\(\uparrow\)  & geo err\(\downarrow\) \\
        \midrule
        \rowcolor{MistyRose} RICH~\cite{huang2022capturing} & 0.6823           & \textbf{0.7427} & 0.6823          & 10.27 \\
        \rowcolor{MistyRose} R+T (2:1)                      & 0.7087           & 0.7370          & \textbf{0.6927} & 9.593 \\
        \rowcolor{MistyRose} R+T (1:1)                      & \textbf{0.7137}  & 0.7286          & 0.6923          & \textbf{9.459} \\
        \midrule
        \rowcolor{LightCyan1} DAMON~\cite{tripathi2023deco}    & 0.6388           & 0.5232          & 0.5115          & 25.06 \\
        \rowcolor{LightCyan1} D+T (2:1)                        & 0.6472           & \textbf{0.5237} & \textbf{0.5148} & 21.54 \\
        \rowcolor{LightCyan1} D+T (1:1)                        & \textbf{0.6701}  & 0.4806          & 0.4972          & \textbf{18.87} \\
        \bottomrule
    \end{tabular}
\end{table}
\vspace{-10pt}

\section{Conclusion}

We introduce \dataset, a large-scale mocap dataset, alongside a novel motion synthesis method, addressing scalability, data quality, and advanced motion synthesis challenges in \ac{hsi} modeling. As the most comprehensive dataset in its category, \dataset encompasses diverse human interactions with dynamic and articulated objects within 100 indoor scenes. Our diffusion-based autoregressive motion synthesis method, leveraging \dataset, is capable of real-time generation of \ac{hsi} sequences of arbitrary length. Experimental results indicate that the motions generated by our method closely mirror the quality of the original \mocap data.

\paragraph{Limitation}

A notable limitation of our method is its inability to generate human-object interaction behaviors beyond those in the training set. For example, it struggles to create realistic motions for unfamiliar actions like climbing off a table, leading to potentially unrealistic interactions or object intersections.

\paragraph{Acknowledgment}

The authors would like to thank NVIDIA for their generous support of GPUs and hardware. This work is supported in part by the National Science and Technology Major Project (2022ZD0114900) and the Beijing Nova Program.

{
    \small
    \bibliographystyle{ieeenat_fullname}
    \bibliography{reference_header,reference}
}

\clearpage
\appendix
\renewcommand\thefigure{A\arabic{figure}}
\setcounter{figure}{0}
\renewcommand\thetable{A\arabic{table}}
\setcounter{table}{0}
\renewcommand\theequation{A\arabic{equation}}
\setcounter{equation}{0}
\pagenumbering{arabic}
\renewcommand*{\thepage}{A\arabic{page}}
\setcounter{footnote}{0}

\section{Additional Details of Experiments}

This section offers a detailed overview of our experimental setup, including the implementation specifics of our method and the necessary adaptations made to baseline methods to ensure fair comparisons. We also elaborate on how our sampling strategy enables real-time control over character motion. For additional qualitative insights and extensive zero-shot transfer experiments, we direct readers to the accompanying \textit{Supplementary Video}.

\subsection{Experiment Settings}

Our experimental setup for motion synthesis methods includes two distinct settings: \textit{static} and \textit{dynamic}.
In the \textit{static} setting, we focus on evaluating the quality of locomotion and scene-level interactions. Each test scene features five predefined pairs of start and end points, given as $(x, y)$ coordinates in a z-up world coordinate system. These points, often located on furniture like chairs, test the method's ability to produce scene-appropriate motions. For complete trajectories, we generate midpoints using a generative A* path planning method, following \citet{wang2022towards}.

The \textit{dynamic} setting involves five pairs of object and human starting locations, accompanied by a trajectory leading towards the object. Each method is tasked with creating five unique motion variations that both approach and interact with the object, conforming to a designated action type.

\subsection{Implementation Details}

Our motion generation model utilizes a DDPM architecture with a linear variance scheduler, conditioned on scene and action embeddings. Following \citet{guo2022generating}, we implement a Transformer encoder as the UNet structure with an embedding dimensionality of 512 for projecting body joint locations into a high-dimensional space. The Transformer consists of 6 layers, 16 heads in the multi-head attention, and uses a 0.1 dropout rate. The intermediate feedforward network's dimensionality is set to 1024. Both scene and action encoders are Transformer-based with 6 layers and 16 heads, producing 512-dimensional embeddings. These embeddings are added to the first token of the motion generation transformer, which contains timestep information. The Huber loss is used to gauge the accuracy of predicted noise.

For converting joint locations to SMPL-X parameterized meshes, a pre-trained 4-layer MLP predicts coarse SMPL-X parameters, with a 6D representation for rotation~\cite{zhou2019continuity}. The MLP inputs three consecutive frames and outputs parameters for the middle frame. Edge cases use duplicated middle frames for input. An optimization process refines body poses using gradient descent on the L2 error between joint locations from the model and predicted SMPL-X parameters, ensuring accurate body pose representation.

Training with the Adam optimizer (learning rate 1e-4, batch size 512) on four NVIDIA A800 GPUs, our method takes 48 hours to train for 500k steps on \dataset.

\subsection{Adaption of Baselines}

We adapted baseline methods for a fair comparison in our autoregressive long-term motion generation framework. For \citet{wang2021synthesizing}, two milestone poses are set as the transition and subgoal at the start and end of each episode, with in-between motions generated using their original cVAE model. Methods like \citet{huang2023sceneDiffuser} and \citet{karunratanakul2023GMD}, not initially designed for long-term synthesis, were modified to incorporate $\mathbf{M}_{goal}$ and $\mathbf{M}_{trans}$ in the sampling stage, maintaining their original sampling strategies. In \textit{dynamic} experiments involving dynamic objects, we adapted models like GOAL~\cite{taheri2021Goal} to encompass both reaching and grasping actions, while preserving their original training pipeline.

\subsection{Human Study}\label{sec:supp:human-study}

We conducted human studies with 31 participants (17 males and 14 females) in a controlled environment, ensuring no communication among them. For each baseline and ablated version of our method, we generated 4 human motions aligned with the respective test settings (dynamic or non-dynamic). These motions were applied to the SMPL-X body mesh and rendered in the same scene from \dataset-test for horizontal comparison. Alongside these synthesized motions, one MoCap motion from \dataset-test in the same scene was also included, with careful rendering to minimize visual obstructions.

To qualitatively compare our method with SAMP~\cite{hassan2021stochastic}, DIMOS~\cite{zhao2023synthesizing}, LAMA~\cite{lee2023locomotion}, and \citet{wang2022towards}, we replicated their demonstrations using our model trained on \dataset. This involved setting similar subgoals to duplicate the trajectories. Participants were shown side-by-side comparisons of motions synthesized by our method and the baseline methods' demos and then asked to choose the more natural-looking output. The frequency of our method being preferred is reflected in the Success Rate of Discrimination (SucRateDis) reported in \cref{tab:human_study}.

\begin{figure*}[t!]
    \centering
    \begin{subfigure}[b]{0.5\linewidth}
        \centering
        \includegraphics[width=\linewidth]{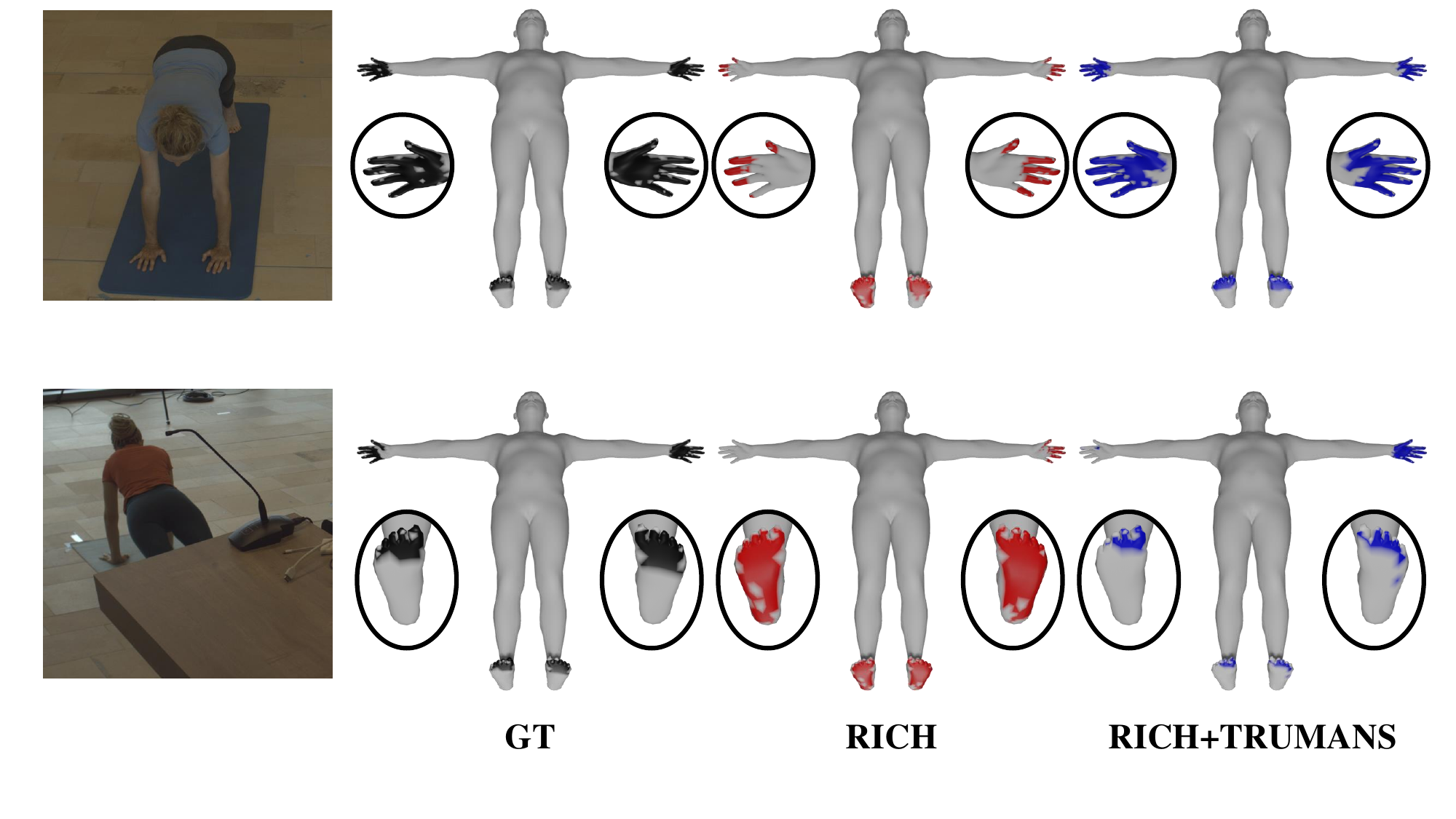}
        \caption{BSTRO~\cite{huang2022capturing} trained on RICH~\cite{huang2022capturing} combined with \dataset.}
    \end{subfigure}%
    \begin{subfigure}[b]{0.5\textwidth}
        \centering
        \includegraphics[width=\linewidth]{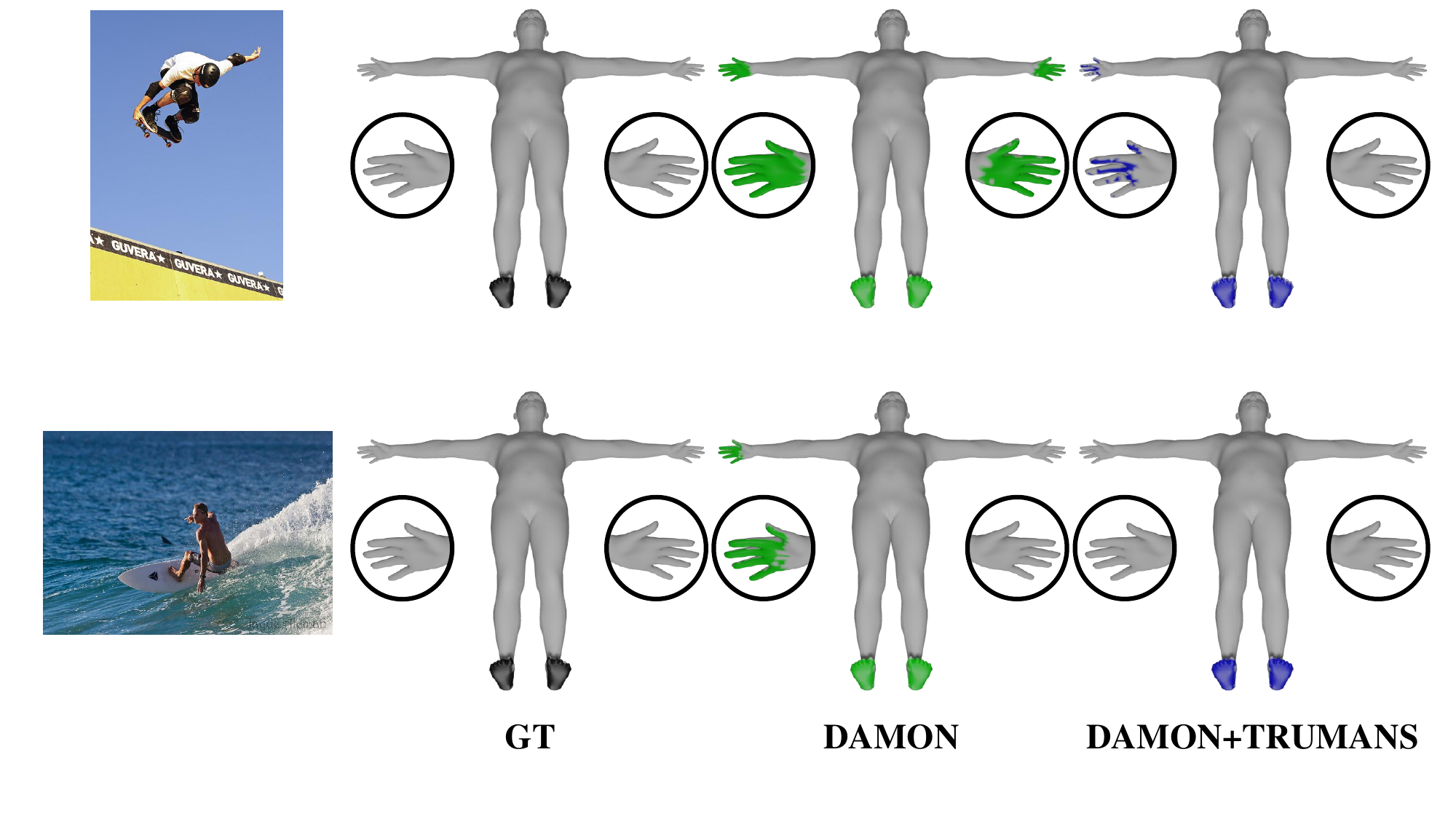}
        \caption{DECO~\cite{tripathi2023deco} trained on DAMON~\cite{tripathi2023deco} combined with \dataset.}
    \end{subfigure}
    \caption{\textbf{Additional qualitative results of 3D contact estimation.}}
    \label{fig:supp:contact}
\end{figure*}

\begin{table}[ht!]
    \centering
    \small
    \setlength{\tabcolsep}{4pt}
    \caption{\textbf{Human study results of comparisons between our method with recent work.} The Success Rate of Discrimination (SucRateDis), indicating the frequency at which our method is selected as the superior one, is reported.}
    \label{tab:human_study}
    \begin{tabular}{lc}
        \toprule  
        Method                            & Success Rate of Discrimination (\%) \\
        \midrule 
        SAMP~\cite{hassan2021stochastic}  & 100 \\
        \citet{wang2022towards} & 100 \\
        LAMA~\cite{lee2023locomotion}     & 80.6 \\
        DIMOS~\cite{zhao2023synthesizing} & 64.5 \\
        \bottomrule
    \end{tabular}
\end{table}

\subsection{Image-based Tasks}\label{sec:supp:image-based-tasks}

This section details additional qualitative and quantitative results for image-based tasks using the rendered images and annotations from \dataset.

\paragraph{3D Human Mesh Estimation}

To assess the impact of integrating \dataset into training, we use two additional methods, I2L~\cite{moon2020i2l} and SGRE~\cite{wang20233d}, on the 3DPW test set. These methods are trained either solely on 3DPW or on a combination of 3DPW and \dataset at varying ratios. As indicated in \cref{tab:supp:mesh_i2l,tab:supp:mesh_SGRE}, incorporating our synthetic data with the real training set markedly enhances performance.

\begin{table}[ht!]
    \centering
    \small
    \setlength{\tabcolsep}{4pt}
    \caption{\textbf{Performance of I2L~\cite{moon2020i2l} in 3D human mesh estimation trained on 3DPW~\cite{von2018recovering} combined with \dataset in different ratios.}}
    \label{tab:supp:mesh_i2l}
    \begin{tabular}{lccc}
        \toprule
        Training Data                 & MPVE\(\downarrow\) & MPJPE\(\downarrow\) & PA-MPJPE\(\downarrow\) \\
        \midrule
        3DPW~\cite{von2018recovering} & 186.9              & 160.4               & 90.2                   \\
        3DPW+T (2:1)                  & 133.2              & 116.5               & 69.1                   \\
        3DPW+T (1:1)                  & \textbf{126.1}     & \textbf{110.2}      & \textbf{66.2}          \\
        \bottomrule 
    \end{tabular}
\end{table}

\begin{table}[ht!]
    \centering
    \small
    \setlength{\tabcolsep}{4pt}
    \caption{\textbf{Performance of SGRE~\cite{wang20233d} in 3D human mesh estimation trained on 3DPW~\cite{von2018recovering} combined with \dataset in different ratios.}}
    \label{tab:supp:mesh_SGRE}
    \begin{tabular}{lccc}
        \toprule
        Training Data                 & MPVE\(\downarrow\) & MPJPE\(\downarrow\) & PA-MPJPE\(\downarrow\) \\
        \midrule
        3DPW~\cite{von2018recovering} & 257.0              & 223.0               & 110.6                  \\
        3DPW+T (2:1)                  & 240.6              & 207.2               & 113.5                  \\
        3DPW+T (1:1)                  & \textbf{138.0}     & \textbf{117.5}      & \textbf{80.3}          \\
        \bottomrule 
    \end{tabular}
\end{table}

\paragraph{3D Contact Estimation}

Qualitative results of baseline methods trained with and without \dataset are presented in \cref{fig:supp:contact}. These results demonstrate that the incorporation of \dataset in training enhances the precision of contact prediction.

\section{Additional Details of \texttt{TRUMANS}}

\subsection{Additional Details of Dataset Comparison}

In our dataset comparison presented in \cref{tab:datasets}, we have categorized similar objects into common types for a more equitable comparison. For SAMP~\cite{hassan2021stochastic}, their seven reported objects, including various chairs and a table, are grouped into three types: ``sofa,'' ``chair,'' and ``table.'' BEHAVE~\cite{bhatnagar2022behave}, with a list of 20 distinct items, is classified into 14 object types, consolidating similar items like chairs and tables. Similarly, iReplica~\cite{guzov23ireplica}'s report of 9 objects is condensed into 5 classes.

Additionally, for iReplica, we have combined data from their two datasets, EgoHOI and H-contact, for simplicity. EgoHOI contributes 0.25 hours of \ac{hsi} data with ego-view and multiview RGBD videos, while H-contact adds 0.5 hours of \ac{hsi} data featuring per-frame hand-object contact.

\subsection{Dataset Splits}

\dataset is divided into training, validation, and test sets, with scenes 1 to 70 for training, 71 to 80 for validation, and 91 to 100 for testing. This distribution creates a split ratio of approximately 7:1:2 across all data frames for the respective sets.

\subsection{Object Types}

\dataset includes 20 types of objects commonly found in indoor scenes, categorized as either \articulated (articulated) or \rigid (rigid). The list with descriptions is as follows:
\begin{itemize}[leftmargin=*,noitemsep,nolistsep]
    \item Articulated chair: \articulated, including gaming and office chairs.
    \item Rigid chair: \rigid, encompasses chairs with/without armrests and stools.
    \item Table: \rigid, available in round and square shapes.
    \item Sofa: \rigid, varieties like single-seaters and couches.
    \item Bed: \rigid.
    \item Book: \articulated.
    \item Pen: \rigid.
    \item Phone: \rigid.
    \item Mouse: \rigid.
    \item Keyboard: \rigid.
    \item Handbag: \rigid.
    \item Vase: \rigid.
    \item Cup: \rigid.
    \item Bottle: \rigid.
    \item Laptop: \articulated.
    \item Oven: \articulated.
    \item Drawer: \articulated.
    \item Cabinet: \articulated.
    \item Microwave: \articulated.
    \item Door: \articulated.
\end{itemize}

\subsection{Capture Pipeline}\label{sec:supp:capture}

\paragraph{Aligning virtual and real environments}

The alignment between virtual and real environments is crucial to ensure the plausibility of actions projected into the virtual world. Our process starts with manually selecting scenes and objects from the 3D-FRONT~\cite{fu20213dfront} dataset and BlenderKit~\cite{blenderkit}, prioritizing scenes with numerous interactable objects that fit within our motion capture area. Manual alignment is performed to match these virtual scenes with real-world counterparts. For example, a digital sofa may be replicated with a real sofa or chairs arranged to mimic its shape. When digital and physical shapes do not match perfectly, we modify the digital asset, such as scaling objects or editing meshes, or adjust our real-world setups, like placing a mat on a chair to simulate a higher seat.

To align digital characters with our real actors, we start by exporting a human armature matching the actor's bone lengths using the VICON Shogun system~\cite{shogun}. Then, in Character Creator 4~\cite{CharacterCreator4}, we adjust sliders to create digital humans mirroring the actors' bone lengths. These digital characters are re-imported into Shogun for real-time \ac{ik} retargeting, ensuring accurate character poses for our digital humans.

\paragraph{Object placeholders}

Our dataset addresses the limitations of previous datasets related to object visibility and complex \acp{hsi} in clustered scenes. We use an optical \mocap system with object placeholders designed for light transmission, enabling accurate capture even in complex \acp{hsi}. For example, an actor seated behind a transparent acrylic table allows for precise leg tracking.

\paragraph{Real-time data quality inspection}

Data recording is monitored in real-time to ensure quality. Inspectors watch the digital world and human avatars on screens during capture, filtering out obvious errors like untracked markers or jittering in \ac{ik} solving. This real-time inspection aids in maintaining high data quality.

\subsection{Motion Augmentation Implementation Details}\label{sec:supp:nsm_detail}

This section delves into the specifics of our motion augmentation pipeline, illustrating the need for our refinement processes with examples and establishing theoretical bounds for the smoothness of \ac{ik} target trajectories.

In our augmentation process, we first identify contact points between human joints and object meshes in a given motion sequence. For example, if joint $J_1$ is in contact with an object mesh at point $\mathbf{v}_m$ at time $T_1$, and we subsequently alter the object's shape or replace it, the new corresponding point becomes $\mathbf{v}_m^\prime$. To preserve the interaction, the joint's target location $\mathbf{l}^\prime$ must be adjusted accordingly to maintain the contact point:
\begin{equation}
    \mathbf{l}_{T_1}^\prime - \mathbf{v}_m^\prime = \mathbf{l}_{T_1} - \mathbf{v}_m,
\end{equation}
such that
\begin{equation}
    \mathbf{l}_{T_1}^\prime = \mathbf{l}_{T_1} + \mathbf{v}_m^\prime - \mathbf{v}_m.
\end{equation}
The offset $\mathbf{v}_1 = \mathbf{v}_m^\prime - \mathbf{v}_m$ represents the change in joint position resulting from the object's shape variance. To address the abrupt trajectory change, we implement a smoothing process for the pose trajectory. Within a defined proximity window $W$, we apply this offset with a linearly decreasing norm to ensure a smoother transition in the joint's trajectory:
\begin{equation}
    \mathbf{v}_{1t} = (1 - \frac{|t-T_1|}{W})\mathbf{v}_1.
\end{equation}
This offset application ensures a seamless blend from the original trajectory to the augmented one. To validate the smoothness of this trajectory, we establish a bound on the difference in bone target positions between consecutive frames. The notation $ | \cdot | $ is used to denote the absolute value for scalars and the 2-norm for vectors, aiding in quantifying these trajectory modifications.
\begin{equation}
    \begin{aligned}
             | \mathbf{l}_{t+1}^{\prime} - \mathbf{l}_{t}^{\prime} | 
         & = | \mathbf{l}_{t+1} + \mathbf{v}_{t+1} - \mathbf{l}_t - \mathbf{v}_t| \\
         &\leq  | \mathbf{l}_{t+1} - \mathbf{l}_t| + | \mathbf{v}_{t+1}  - \mathbf{v}_t|  \\
         &\leq | \mathbf{l}_{t+1} - \mathbf{l}_t| + \frac{|\mathbf{v}|}{W}.
     \end{aligned}
     \label{eq:bound_simple}
\end{equation}
When the smoothing window length $W$ is set appropriately, such as 30 in our implementation, and the norms of the offset vectors are within practical limits (dictated by reasonable object variation scales and stable CCD \ac{ik} results), the updated \ac{ik} target trajectories exhibit sufficient smoothness.

However, this \ac{ik} solving approach is limited to simpler scenarios involving interaction with a single object. In more complex situations, typical of our dataset, we encounter additional challenges. For instance, consider an offset $\mathbf{v}_1$ introduced at time $T_1$ for bone $J_1$ due to object augmentation, with this contact ending at $T_2$. At $T_2$, another bone $J_2$ enters into contact, necessitating a new \ac{ik} target. Post-\ac{ik} process, bones without specific targets, including $J_1$, may shift from their original positions. We denote the deviation for bone $J_1$ as $\mathbf{v}_2$. To mitigate this deviation, we employ a similar offsetting technique by blending $\mathbf{v}_1$ and $\mathbf{v}_2$. For each time $t$ within this window of length $W$, the bone is assigned an offset vector that is a weighted mean of these two offsets, calculated using vector norms as weights:
\begin{equation}
    \text{offset} = \frac{(|\mathbf{v}_{1t}| \mathbf{v}_{1t} + |\mathbf{v}_{2t}|\mathbf{v}_{2t})}{|\mathbf{v}_{1t}|+|\mathbf{v}_{2t}|},
\end{equation}
where 
\begin{equation}
    \begin{aligned}
        \mathbf{v_{1t}} &= (1 - \frac{|t-T_1|}{W})\mathbf{v}_1,\\ 
        \mathbf{v_{2t}} &= (1 - \frac{|t-T_2|}{W})\mathbf{v}_2.
    \end{aligned}
\end{equation}

\begin{figure*}[t!]
    \centering
    \includegraphics[width=0.86\linewidth]{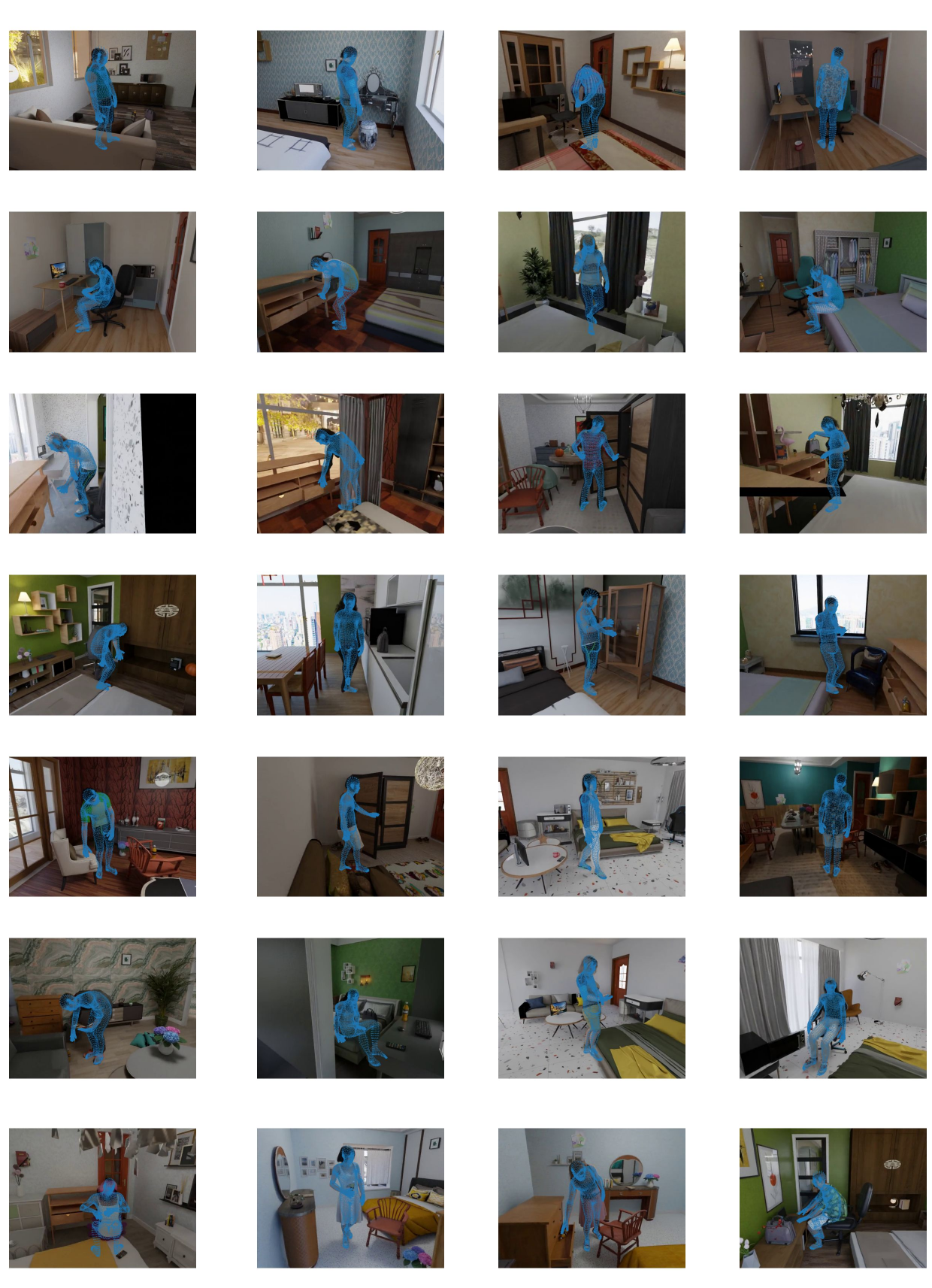}
    \caption{\textbf{Examples of SMPL-X annotations in \dataset.}}
    \label{fig:supp:smplx-vis}
\end{figure*}

\begin{figure*}[t!]
    \centering
    \includegraphics[width=\linewidth]{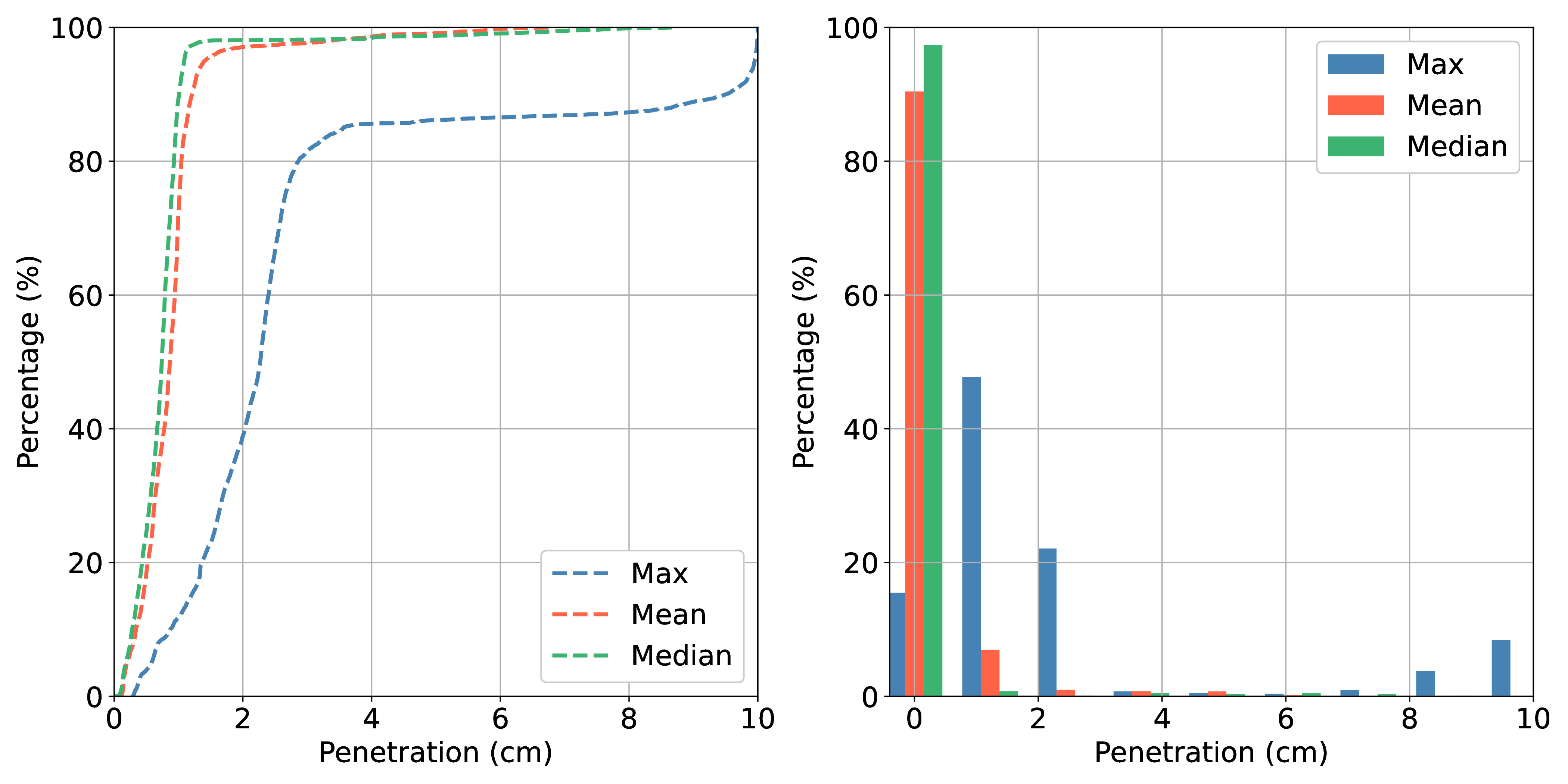}
    \caption{\textbf{Penetration statistics in \dataset.} This analysis covers maximum, mean, and median penetration distances per vertex per frame. The left graph displays the proportion of frames with penetration below various thresholds (X-axis), and the right bar plot categorizes frames by specific penetration distance ranges (X-axis). Notably, in more than 95\% of frames, both the mean and median penetration distances stay below 2cm.}
    \label{fig:supp:penetration}
\end{figure*}

By integrating these two stages of linear blending of offset vectors, we achieve smooth trajectories for \ac{ik} targets. As outlined earlier, akin to the approach in \cref{eq:bound_simple}, we analyze the joint target differences between consecutive frames to further substantiate the smoothness of our approach:
\begin{align}
    & | \mathbf{l}_{t+1}^{\prime} - \mathbf{l}_t^{\prime}| \nonumber\\
    =\quad&\bigg\vert \mathbf{l}_{t+1} - \mathbf{l}_{t} \nonumber\\
    & + \frac{|\mathbf{v}_{1,t+1}| \mathbf{v}_{1,t+1} + |\mathbf{v}_{2,t+1}| \mathbf{v}_{2,t+1} }{|\mathbf{v}_{1,t+1}| + |\mathbf{v}_{2,t+1}| } \\
    & - \frac{|\mathbf{v}_{1,t}| \mathbf{v}_{1,t} + |\mathbf{v}_{2,t}| \mathbf{v}_{2,t} }{|\mathbf{v}_{1,t}| + |\mathbf{v}_{2,t}| } \bigg\vert \nonumber\\
\leq\quad& | \mathbf{l}_{t+1} - \mathbf{l}_{t} | \\
    & +\bigg\vert \frac{|\mathbf{v}_{1,t+1}| \mathbf{v}_{1,t+1}}{|\mathbf{v}_{1,t+1}| + |\mathbf{v}_{2,t+1}| } - \frac{|\mathbf{v}_{1,t}| \mathbf{v}_{1,t}}{|\mathbf{v}_{1,t}| + |\mathbf{v}_{2,t}| } \bigg\vert \label{nsm_proof_part_v1}\\
    & +\bigg\vert \frac{|\mathbf{v}_{2,t+1}| \mathbf{v}_{2,t+1}}{|\mathbf{v}_{1,t+1}| + |\mathbf{v}_{2,t+1}| } - \frac{|\mathbf{v}_{2,t}| \mathbf{v}_{2,t}}{|\mathbf{v}_{1,t}| + |\mathbf{v}_{2,t}| } \bigg\vert, \label{nsm_proof_part_v2}
\end{align}
where 
\begin{equation}
    |\mathbf{v}_{i,t}| = (1 - |\frac{t-T_i}{W}|) |\mathbf{v}_i|.
\end{equation}
Thus we have
\begin{equation}
    \begin{aligned}
        & \bigg\vert \frac{|\mathbf{v}_{1,t+1}| \mathbf{v}_{1,t+1}}{|\mathbf{v}_{1,t+1}| + |\mathbf{v}_{2,t+1}| } - \frac{|\mathbf{v}_{1,t}| \mathbf{v}_{1,t}}{|\mathbf{v}_{1,t}| + |\mathbf{v}_{2,t}|} \bigg\vert \\
    =\quad& \frac{|\mathbf{v}_{1,t+1}||\mathbf{v}_{1,t}| (\mathbf{v}_{1,t+1} - \mathbf{v}_{1,t})}{(|\mathbf{v}_{1,t+1}| + |\mathbf{v}_{2,t+1}|)(|\mathbf{v}_{1,t}| + |\mathbf{v}_{2,t}|)}\\
        & + \frac{| (|\mathbf{v}_{1,t+1}||\mathbf{v}_{2,t}|\mathbf{v}_{1,t+1} - |\mathbf{v}_{2,t+1}||\mathbf{v}_{1,t}|\mathbf{v}_{1,t}) | }{(|\mathbf{v}_{1,t+1}| + |\mathbf{v}_{2,t+1}|)(|\mathbf{v}_{1,t}| + |\mathbf{v}_{2,t}|)} \\
    \leq\quad& \frac{|\mathbf{v}_{1,t+1}||\mathbf{v}_{1,t}| |\mathbf{v}_{1,t+1} - \mathbf{v}_{1,t}|}{(|\mathbf{v}_{1,t+1}| + |\mathbf{v}_{2,t+1}|)(|\mathbf{v}_{1,t}| + |\mathbf{v}_{2,t}|)}\\
        & + \bigg\vert \frac{|\mathbf{v}_{1,t+1}||\mathbf{v}_{2,t}|\mathbf{v}_{1,t+1} - |\mathbf{v}_{1,t+1}||\mathbf{v}_{2,t}|\mathbf{v}_{1,t}}{(|\mathbf{v}_{1,t+1}| + |\mathbf{v}_{2,t+1}|)(|\mathbf{v}_{1,t}| + |\mathbf{v}_{2,t}|)} \bigg\vert\\
        & + \bigg\vert \frac{|\mathbf{v}_{1,t+1}||\mathbf{v}_{2,t}|\mathbf{v}_{1,t} - |\mathbf{v}_{1,t}||\mathbf{v}_{2,t}|\mathbf{v}_{1,t}}{(|\mathbf{v}_{1,t+1}| + |\mathbf{v}_{2,t+1}|)(|\mathbf{v}_{1,t}| + |\mathbf{v}_{2,t}|)} \bigg\vert\\
        & + \bigg\vert \frac{|\mathbf{v}_{1,t}||\mathbf{v}_{2,t}|\mathbf{v}_{1,t} - |\mathbf{v}_{1,t}||\mathbf{v}_{2,t+1}|\mathbf{v}_{1,t}}{(|\mathbf{v}_{1,t+1}| + |\mathbf{v}_{2,t+1}|)(|\mathbf{v}_{1,t}| + |\mathbf{v}_{2,t}|)} \bigg\vert \\
    <\quad& \frac{|\mathbf{v}_{1,t+1}||\mathbf{v}_{1,t}| |\mathbf{v}_{1,t+1} - \mathbf{v}_{1,t}|}{|\mathbf{v}_{1,t+1}||\mathbf{v}_{1,t}|}\\
        & + \frac{|\mathbf{v}_{1,t+1}||\mathbf{v}_{2,t}|\mathbf{v}_{1,t+1} - \mathbf{v}_{1,t}|}{|\mathbf{v}_{1,t+1}||\mathbf{v}_{2,t}|}\\
        & + \bigg\vert\frac{|\mathbf{v}_{1,t}||\mathbf{v}_{2,t}|(|\mathbf{v}_{1,t+1}| - |\mathbf{v}_{1,t}|)}{|\mathbf{v}_{1,t}||\mathbf{v}_{2,t}|} \bigg\vert\\
        & + \bigg\vert\frac{|\mathbf{v}_{1,t}||\mathbf{v}_{1,t}|(|\mathbf{v}_{2,t+1}| - |\mathbf{v}_{2,t}|)}{|\mathbf{v}_{2,t+1}||\mathbf{v}_{2,t}|} \bigg\vert \\
    =\quad& \frac{3|\mathbf{v}_1| + |\mathbf{v}_2|}{W}.
    \end{aligned}
    \label{nsm_proof_part_v1_result} 
\end{equation}
With the scaling of \cref{nsm_proof_part_v1,nsm_proof_part_v2} now aligned, we substitute these elements to establish a theoretical bound:
\begin{equation}
    \begin{aligned}
            & | \mathbf{l}_{t+1}^{\prime} - \mathbf{l}_t^{\prime}| \\
        <\quad& | \mathbf{l}_{t+1} - \mathbf{l}_{t}|
         +  \frac{3|\mathbf{v}_1| + |\mathbf{v}_2|}{W}
         + \frac{|\mathbf{v}_1| + 3|\mathbf{v}_2|}{W} \\
        =\quad& | \mathbf{l}_{t+1} - \mathbf{l}_{t} | + \frac{4}{W} ( |\mathbf{v}_1| + |\mathbf{v}_2|),
    \end{aligned}
\end{equation}
which ensures that our target trajectories are smooth.

\begin{figure*}[t!]
    \centering
    \includegraphics[width=0.9\linewidth]{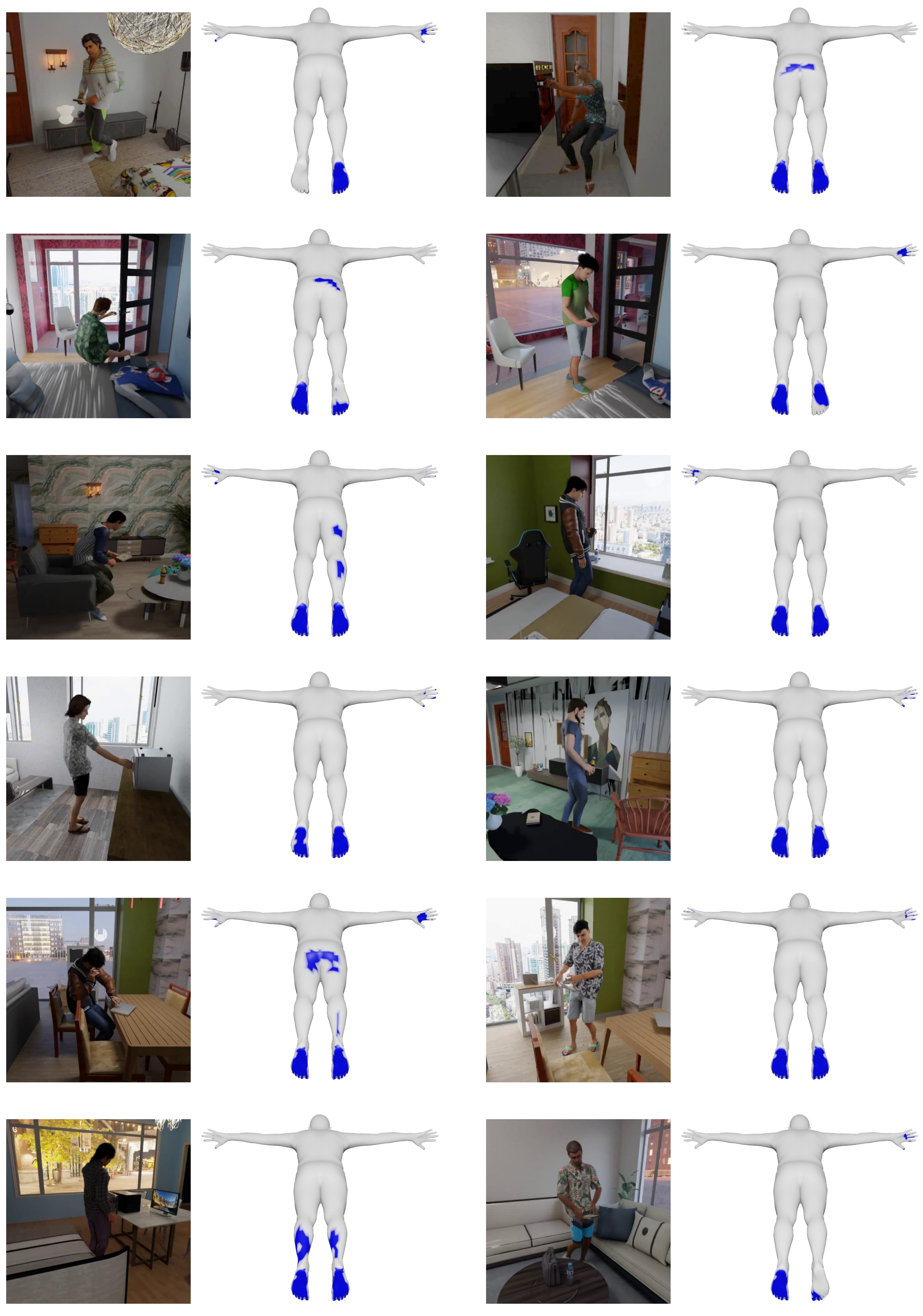}
    \caption{\textbf{Examples of contact annotations in \dataset.}}
    \label{fig:supp:contact-vis}
\end{figure*}

\subsection{Annotations}

\paragraph{Human Motion}

The motion data captured from the \vicon system, initially in the form of pose sequences of a custom armature, is converted into the SMPL-X format~\cite{pavlakos2019expressive} using a vertex-to-vertex optimization method. This ensures accurate and smooth SMPL-X representations; please refer to \cref{fig:supp:smplx-vis} for examples. The conversion process involves the following steps:
\begin{enumerate}[leftmargin=*,noitemsep,nolistsep]
    \item Vertices on the SMPL-X mesh are manually selected and paired with the closest vertices on our custom mesh.
    \item A loss function, defined as the Mean Squared Error (MSE) between paired vertex locations, is minimized using the Adam optimizer to refine SMPL-X parameters until convergence.
    \item Inaccuracies in the SMPL-X mesh are manually corrected by adjusting bone rotations.
\end{enumerate}
After aligning the SMPL-X mesh with our custom mesh, we record the mapping for future use.

In the second phase, we enhance the custom motion data by adding interpolated frames from the T-pose to the start pose. This ensures a smooth transition for each bone in the sequence.

Finally, we optimize the SMPL-X parameters frame by frame, starting from the pose established in the first phase. We first refine the body shape parameters and then adjust the pose parameters, including global translation. The optimization of each frame starts from the pose of the previous frame and continues until convergence. This method relies on minimal mesh changes between frames, supported by our high-quality motion data. A typical MSE value at convergence ranges between 5e-5 and 1e-4, indicating an average point distance of less than 1cm. 

\paragraph{Contact}

Following the fitting of SMPL-X meshes, we compute per-vertex contact annotations. The contact for each human mesh vertex is determined based on its proximity and orientation relative to scene or object meshes. A vertex is deemed in contact if it fulfills either of the following conditions: (i) it resides inside an object's mesh, or (ii) while outside an object's mesh, it is within a specified threshold distance and the angle between its normal and the vector pointing towards the object is under 60 degrees. The latter criterion is particularly vital for accurate contact annotation, as it prevents misidentification in scenarios like a hand holding a bottle. Penetration statistics, as detailed in \cref{fig:supp:penetration}, reveal that in over 95\% of the frames, both the mean and median penetration distances remain below 2cm. For examples of contact annotation, please refer to \cref{fig:supp:contact-vis}.

\paragraph{Objects}

In \dataset, we include the watertight mesh for all objects and their 6D poses for each frame. For articulated objects, part-level annotations are provided along with the URDF (Unified Robot Description Format) files to represent their kinematic structure.

\paragraph{Actions}

For every sequence within \dataset, multi-hot action labels are assigned on a frame-by-frame basis. This approach allows for the representation of multiple concurrent actions in a single frame, with each action identified by a distinct label.

\subsection{Video Rendering}

To enhance video diversity and accurately capture \ac{hsi} details, we developed an adaptive camera tracking algorithm that maintains footage smoothness and consistency. The camera, set at a constant height of 1.4 meters, moves within a 2-meter radius around the human, horizontally oriented towards the body. The camera's pose is exclusively determined by its rotation around the z-axis in the human coordinate system.

We set keyframes at intervals of 30 frames. For each keyframe, 20 camera proposals adhering to the constraints are pre-defined and evenly distributed around the ring. To identify active hand interactions, we calculate the minimum distance between hand joints and dynamic objects. If this distance exceeds 20 centimeters, we default to tracking the right hand. For the identified interacting hand, rays emitted from each camera proposal towards the hand joints help measure visibility. A joint is considered visible to a camera if the intersection point with the scene's mesh can be projected onto the camera's imaging plane, and the distance to the joint is less than 10 centimeters. The number of visible joints determines whether a camera effectively captures the interaction. The visibility threshold for different keyframes is dynamically adjusted to ensure at least one camera captures the interaction, except when all joints are invisible.

After assessing the interaction capture capability of the 20 cameras at each keyframe, dynamic programming is used to select the optimal keyframe camera pose sequence that minimizes total rotation and maximizes interaction coverage. ``Camera pose'' here specifically refers to rotation about the z-axis. Camera poses for frames between keyframes are interpolated using cubic spline interpolation of the rotation angles.

\end{document}